# Improved Memory-Bounded Dynamic Programming for Decentralized POMDPs

**Sven Seuken**
School of Engineering and Applied Sciences
Harvard University
Cambridge, MA 02138
seuken@eecs.harvard.edu

**Shlomo Zilberstein**
Department of Computer Science
University of Massachusetts
Amherst, MA 01003
shlomo@cs.umass.edu

## Abstract

Memory-Bounded Dynamic Programming (MBDP) has proved extremely effective in solving decentralized POMDPs with large horizons. We generalize the algorithm and improve its scalability by reducing the complexity with respect to the number of observations from exponential to polynomial. We derive error bounds on solution quality with respect to this new approximation and analyze the convergence behavior. To evaluate the effectiveness of the improvements, we introduce a new, larger benchmark problem. Experimental results show that despite the high complexity of decentralized POMDPs, scalable solution techniques such as MBDP perform surprisingly well.

## 1 Introduction

The *Markov decision process* (MDP) and its partially observable counterpart (POMDP) have proved very useful for planning and learning under uncertainty. Decentralized POMDPs (DEC-POMDPs) offer a natural extension of these frameworks for cooperative multi-agent settings. They capture effectively situations in which agents have different partial knowledge about the state of the environment and the other agents. Many decentralized decision problems in real life, – such as multi-robot coordination, manufacturing, information gathering and load balancing – can be modeled as DEC-POMDPs. Solving finite-horizon DEC-POMDPs is NEXP-complete [Bernstein *et al.*, 2000] and even $\varepsilon$-approximations are hard [Rabinovich *et al.*, 2003]. Thus, optimal algorithms have mostly theoretical significance. A detailed survey of existing formal models, complexity results and planning algorithms is available in [Seuken and Zilberstein, 2005].

Over the last five years, researchers have introduced several algorithms for solving DEC-POMDPs approximately. Unfortunately, these algorithms cannot solve problems with horizons larger than 10. One exception is the Memory-Bounded Dynamic Programming (MBDP) algorithm [Seuken and Zilberstein, 2007]. MBDP's runtime grows polynomially with the horizon, as opposed to the double exponential growth of exact algorithms. As a result, it can solve problems with horizons that are multiple orders of magnitudes larger than what was previously possible (e.g. horizon 100,000 for the multi-agent tiger problem).

Unfortunately, MBDP's memory requirements still grow exponentially with the size of the observation set. Even for very small problems this can be prohibitive. The main contribution of this paper is an improved version of MBDP that is based on the insight that in most belief states, only a few observations are likely or even possible. Optimizing a policy for unlikely observations at a specific belief point is unnecessary. Thus the exponential growth with respect to the observation space could be avoided without sacrificing solution quality too much. The details of the improved MBDP algorithm are presented in Section 4. Section 5 provides a theoretical analysis of the improved algorithm, establishes an error bound for the new approximation technique and analyzes the convergence behavior. To test the algorithm, we introduce a new, harder benchmark problem in Section 6. Despite the high worst-case complexity of DEC-POMDPs, the improved MBDP algorithm can find quickly good solutions. We conclude with a discussion of open research questions and promising future research directions.

## 2 Related Work

Since the introduction of the DEC-POMDP model in 2000, researchers have developed a wide range of optimal and approximate algorithms. We focus here on finite-horizon problems; infinite-horizon problems often require different solution representation and different techniques. One of the first approximate algorithms is the Joint Equilibrium-based Search for Poli-cies. This algorithm does not search for a globally optimal solution, but instead aims for local optimality [Nair et al., 2003]. A nice feature of the algorithm is its focus on *reachable belief states*. The Point-Based Dynamic Programming (PBDP) algorithm [Szer and Charpillet, 2006] extends this idea further and computes policies based on a subset of the reachable belief states. It is also the most closely related technique to the original MBDP algorithm, but unlike MBDP it is not memory-bounded. If a DEC-POMDP is treated as a partially observable stochastic game (POSG) with common payoffs, some game-theoretic techniques can be exploited. For example, POSGs can be approximated by a series of smaller Bayesian games [Emery-Montemerlo et al., 2004]. All of these algorithms improve upon the performance of the optimal planning algorithms, but true scalability remains a challenge. Specifically, with the exception of MBDP, these algorithms cannot solve existing benchmark problems with horizons beyond 10. MBDP was the first to overcome this particular barrier, quickly finding near-optimal solutions for problems with horizons that are several orders of magnitude larger.

## 3 Solving Decentralized POMDPs

We formalize the problem using the DEC-POMDP model [Bernstein et al., 2000]. Our results, however, apply to equivalent models such as MTDP or COM-MTDP [Pynadath and Tambe, 2002].

**Definition 1 (DEC-POMDP)** A finite-horizon decentralized partially-observable Markov decision process is a tuple $\langle I, S, b^0, \{A_i\}, P, \{\Omega_i\}, O, R, T \rangle$ s.t.

- $I$ is a finite set of agents indexed 1,...,n.
- $S$ is a finite set of states, i.e. the state space.
- $b^0 \in \Delta S$ is the initial belief state (state distribution).
- $A_i$ is a finite set of actions available to agent $i$ and $\vec{A} = \otimes_{i \in I} A_i$ is the set of joint actions, where $\vec{a} = \langle a_1, ..., a_n \rangle$ denotes a joint action.
- $P$ is a Markovian transition probability table. $P(s'|s, \vec{a})$ is the probability that taking joint action $\vec{a}$ in state $s$ results in state $s'$.
- $\Omega_i$ is a finite set of observations available to agent $i$ and $\vec{\Omega} = \otimes_{i \in I} \Omega_i$ is the set of joint observations, where $\vec{o} = \langle o_1, ..., o_n \rangle$ denotes a joint observation.
- $O$ is a table of observation probabilities. $O(\vec{o}|\vec{a}, s)$ is the probability of joint observation $\vec{o}$ given that joint action $\vec{a}$ was taken in state $s$.
- $R : S \times \vec{A} \times S \to \Re$ is a reward function. $R(s, \vec{a}, s')$ is the reward obtained from transitioning to state $s'$ after taking joint action $\vec{a}$ in state $s$.
- $T$ denotes the total number of time steps.

Because the underlying system state $s$ of a DEC-POMDP is not available to the agents during execution time, they must base their actions on beliefs about the current situation. In a single-agent setting, the *belief state* or *belief point* $b$, a distribution over states, is sufficient for optimal action selection. In a distributed setting, each agent must base its actions on a *multi-agent belief state* – a distribution over states and over the other agents' policies. This introduces particular challenges for approximate algorithms which are discussed in detail in Section 7.

**Definition 2 (Policies for a DEC-POMDP)** A **local policy** for agent $i$, $\delta_i$, is a mapping from local histories of observations $\bar{o}_i = o_{i1} \cdots o_{it}$ over $\Omega_i$, to actions in $A_i$. A **joint policy**, $\delta = \langle \delta_1, ..., \delta_n \rangle$, is a tuple of local policies, one for each agent.

Solving a DEC-POMDP means finding a joint policy that maximizes the expected total reward $E[\sum_{t=1}^{T} R(s_{t-1}, \vec{a}, s_t | b^0)]$. A policy for a single agent $i$ can be represented as a decision tree $q_i$, where nodes are labeled with actions and arcs are labeled with observations (a so called *policy tree*). A solution to a DEC-POMDP with horizon $t$ can then be seen as a vector of horizon-$t$ policy trees, a so called *joint policy tree* $\delta^t = (q_1^t, q_2^t, ..., q_n^t)$, one policy tree for each agent, where $q_i^t \in Q_i^t$. The value of a policy tree of a single agent is highly dependent on the policy trees of the other agents and on all agents' beliefs about the current situation. This includes the beliefs about the underlying system state and the beliefs about the policies used by the other agents. Thus, the value of a single agent policy cannot be determined in isolation - only a joint policy with a given start state has a meaningful value. The value of a joint policy given an initial belief state is defined as follows.

**Definition 3 (Value of Joint Policy Tree)** The value of a joint policy $\delta^T$ for an initial belief state $b^0$ is defined as: $V(\delta^T, b^0) = \sum_{s \in S} b^0(s) \cdot V(\delta^T, s)$.

The policy trees for the agents can be constructed in two different ways: top-down or bottom-up. If the goal is an approximate solution, the different characteristics of the construction processes can be exploited.

**Top-down Approach** The first algorithm that used a top-down approach, MAA$^*$, makes use of heuristic search techniques [Szer et al., 2005]. It is an extension of the standard A$^*$ algorithm where each search node contains a joint policy tree. For example, if $\delta^2$ is a horizon-2 joint policy tree, an expansion of the corresponding search node generates all possible joint policy trees of horizon 3 that use the joint policy tree $\delta^2$ for the first two time steps. Using heuristics that are suitable for DEC-POMDPs, some parts of the search tree can be pruned. Unfortunately, the algorithm runs out of time even for small problems with very short horizons, because the search space grows double exponentially with the horizon (see [Seuken and Zilberstein, 2005] for a detailed discussion of this problem).

**Bottom-up Approach: Dynamic Programming**
The first non-trivial algorithm for solving DEC-POMDPs used a bottom-up approach [Hansen *et al.*, 2004]. Policy trees were constructed incrementally, but instead of successively coming closer to the frontiers of the trees, this algorithm starts at the frontiers and works its way up using dynamic programming (DP). A separate set of policy trees is kept for each agent. In the end, the best policy trees are combined to produce the optimal joint policy.

The DP algorithm starts by constructing all possible one-step policies. In each consecutive iteration, the algorithm uses a set of horizon-$t$ policy trees $Q_i^t$ and creates the set $Q_i^{t+1}$ by an *exhaustive backup*. This operation generates every possible depth-$(t+1)$ policy tree that makes a transition, after an action and observation, to the root node of some depth-$t$ policy tree. The total number of complete policy trees for each agent is of the order $\mathcal{O}(|A|^{(|O|^T)})$. This double exponential blow-up is why a naive algorithm would quickly run out of memory. To alleviate this problem, the algorithm uses *iterated elimination of dominated policies* after each backup. This technique significantly reduces the number of policy trees kept in memory without sacrificing the value of the optimal joint policy. Unfortunately, even with this pruning technique, the number of policy trees still grows quickly and the algorithm runs out of memory.

An analysis of the construction process reveals that most of the policies kept in memory are useless because a policy tree can only be eliminated if it is dominated for *every* belief state. But for many DEC-POMDPs, only a small subset of the belief space is actually reachable. Furthermore a policy tree can only be eliminated if it is dominated for every possible belief over the other agents' policies. But obviously, during the construction process, the other agents also maintain a large set of policy trees that will eventually prove to be useless. Unfortunately, with a bottom-up approach, these drawbacks of the pruning process cannot be avoided. Before the algorithm reaches the roots of the final policy trees, it cannot predict which beliefs about the state and about the other agents' policies will eventually be useful. This observation was the main insight used for developing the MBDP algorithm, which combines the bottom-up and top-down approaches: Top-down heuristics identify relevant belief states for which the DP algorithm can evaluate the bottom-up policy trees and select the best joint policy.

## 4 The Improved MBDP Algorithm

Even though the agents do not have access to the belief state during execution time, it can be used to evaluate the bottom-up policy trees computed by the DP algorithm. A set of belief states can be computed using multiple top-down heuristics – efficient algorithms that find useful top-down policies. Once a top-down heuristic policy is generated, the most likely belief states can be computed. In [Seuken and Zilberstein, 2007] we describe a portfolio of heuristics suitable for DEC-POMDPs. In practice, the MDP-heuristic (revealing the underlying system state after each time step) and a random-policy heuristic have proven useful. The usefulness of the heuristics and, more importantly, the computed belief states are highly dependent on the specific problem. But once the algorithm has computed a complete solution, we have a joint policy that definitely leads to relevant belief states when used as a heuristic. This is exactly the idea of *Recursive MBDP*. The algorithm can be applied recursively with an arbitrary recursion-depth.

In the original MBDP algorithm the policy trees for each agent are constructed incrementally using the bottom-up DP approach. To avoid the double exponential blow-up, the parameter *maxTrees* is chosen such that a full backup with this number of trees does not exceed the available memory. Every iteration of the algorithm consists of the following steps. First, a full backup of the policies from the last iteration is performed. This creates policy tree sets of the size $|A||maxTrees|^{|O|}$. Next, top-down heuristics are chosen from the portfolio and used to compute a set of belief states. Then, the best policy tree pairs for these belief states are added to the new sets of policy trees. Finally, after the $T$th backup, the best joint policy tree for the start distribution is returned.

### 4.1 Motivation for new Approximation

Although the original MBDP algorithm has a linear time and space complexity with respect to the horizon $T$, it is still exponential in the size of the observation space. This is a severe limitation when tackling just slightly larger problems than the multi-agent broadcast channel problem or the multi-agent tiger problem. Even for a small problem with 2 actions and 5 observations, setting $maxTrees = 5$ would be prohibitive for the algorithm because $(2 \cdot 5^5)^2 = 39,062,500$ policy tree pairs would have to be evaluated. None of the existing algorithms can solve a DEC-POMDP with 2 actions and 5 observations.

Consequently, tackling the size of the observation set is crucial. The key observation is that considering the whole set of possible observations in every belief state and for every possible horizon is not useful and actu-ally not necessary. Consider for example robots navigating in a building. After turning away from the entrance door, the robots are highly unlikely to observe that same door in the next time step. In general, depending on the belief state and the action choice, only a very limited set of observations might be possible.

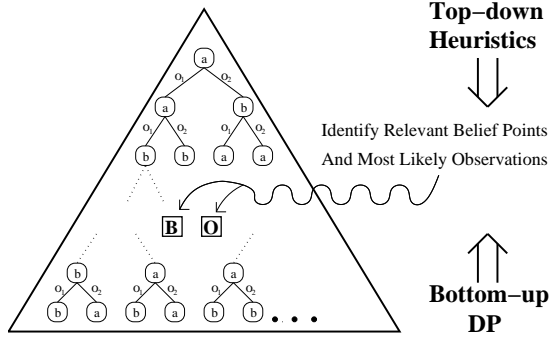

Figure 1: The construction of a single policy tree combining the top-down and the bottom-up approaches.

### 4.2 The Improvement: Partial Backups

In order to select a limited set of useful observations we propose a similar technique used for selecting promising bottom-up policy trees. As before, in step one the algorithm first identifies a set of belief states using the top-down heuristics. For every identified belief state $b^t$ at horizon $t$ the best joint policy is added. Additionally, in step two the set of most likely observations for every agent is identified, bounded by a pre-defined number of observations $maxObs$. More specifically, for the most likely belief state $b^{t-1}$ for horizon $t-1$ and joint action $\vec{a}$ prescribed by the heuristic, the probability $Pr(\vec{o}) = \sum_s b^{t-1}(s) \cdot O(\vec{o}|\vec{a}, s)$ is computed for each joint observation $\vec{o}$. Then all joint observations $\vec{o}$ are ranked according to $Pr(\vec{o})$. This ranking is used to select the $k$ most likely observations for each agent separately, where $k = maxObs$. For the joint policy trees selected in step one the algorithm then performs a *partial backup* with just $k$ observations for every agent, leading to a feasible number of policy trees (with missing branches). Note that although only $k$ observations are used for the partial backup of every agent's policy trees, implicitly $k^n$ joint observations are taken into consideration. After the partial backups the resulting trees are filled up with the missing observation branches using local search (hill climbing: for every missing observation check which of the available subnodes of the next level will lead to the highest expected utility for the given belief state). This concludes a single iteration. Figure 1 illustrates the idea of the improved MBDP algorithm, i.e. the combination of the two approximation techniques. The algorithm can be applied to DEC-POMDPs with an arbitrary number of agents, but to simplify notation, the description shown in Algorithm 1 is for two agents, $i$ and $j$.

---

**Algorithm 1**: The Improved MBDP Algorithm
**begin**
  $maxTrees \leftarrow$ max number of trees before backup
  $maxObs \leftarrow$ max number of observations for backup
  $T \leftarrow$ horizon of the DEC-POMDP
  $H \leftarrow$ pre-compute heuristic policies for each $h \in H$
  $Q_i^1, Q_j^1 \leftarrow$ initialize all 1-step policy trees
  **for** $t=1$ to $T-1$ **do**
    $Sel_i^t, Sel_j^t \leftarrow$ empty
    **for** $k=1$ to $maxTrees$ **do**
      choose $h \in H$ and generate belief state $b^{T-t}$
      **foreach** $q_i \in Q_i^t, q_j \in Q_j^t$ **do**
        evaluate pair $(q_i, q_j)$ with respect to $b^{T-t}$
      add best policy trees to $Sel_i^t$ and $Sel_j^t$
      delete these policy trees from $Q_i^t$ and $Q_j^t$
    choose $h \in H$ and generate belief state $b^{T-t-1}$
    $O_i, O_j \leftarrow maxObs$ most likely obs. for $h(b^{T-t-1})$
    $Q_i^{t+1}, Q_j^{t+1} \leftarrow$ partialBackup($Sel_i^t, Sel_j^t, O_i, O_j$)
    fill up $Q_i^{t+1}$ and $Q_j^{t+1}$ with missing observations
    improve $Q_i^{t+1}$ and $Q_j^{t+1}$ via hill climbing
  select best joint policy tree $\delta^T$ for $b^0$ from $\{Q_i^T, Q_j^T\}$
  return $\delta^T$
**end**

## 5 Theoretical Analysis

The quality of the joint policy trees produced by the improved MBDP algorithm depends on the maximum number of trees per horizon level $maxTrees$ and the maximum number of observations allowed per partial backup $maxObs$. In this section we prove that the error introduced due to the partial backup approximation is bounded and that the bound converges to 0 as $maxObs$ is increased.

### 5.1 Error Bound

The parameter $maxObs$ determines how many observations the improved MBDP algorithm considers in its partial backup for each agent. Taking less than the maximum number of observations into consideration introduces an approximation error. For any belief state $b$ at horizon $t$ and joint action $\vec{a}$, let $\varepsilon^t(b, \vec{a})$ denote the sum of the probabilities of the $m$ most likely joint observations $\vec{O}^*$ in the $(t+1)$th step, where we require that each agent separately has at most $maxObs$ local observations. More formally:

$$\varepsilon^t(b, \vec{a}) = \max_{\vec{O}^* = \{\vec{o}_1, \vec{o}_2, \ldots\}} \sum_{\vec{o}_i \in \vec{O}^*} \sum_{s \in S} b(s) \cdot O(\vec{o}_i | \vec{a}, s)$$

where $\forall$ agents $j$ ($f_j(\vec{o}_i)$ denotes the $j$th element of $\vec{o}_i$)

$$|\{f_j(\vec{o}_1), f_j(\vec{o}_2), \ldots, f_j(\vec{o}_m)\}| \leq maxObs$$

Note that the set of possible belief states at horizon $t$ is determined by the set of possible joint observation sequences. Given observation sequence $\theta$, we can compute $P(s|\theta)$ for all $s$ which gives us a belief state. Thus

we can define:
$$\varepsilon^t(\theta, \vec{a}) = \varepsilon^t(b(\theta), \vec{a})$$
and finally:
$$\varepsilon = \min_{1 \leq t \leq T} \min_{\theta \in \Theta^t} \min_{\vec{a} \in \vec{A}} \varepsilon^t(\theta, \vec{a})$$

Thus, for every possible horizon, every possible joint observation sequence and ever possible joint action the sum of the probabilities of the joint observations used in a partial backup will always be at least $\varepsilon$. This parameter describes the maximum error introduced by the new approximation in the improved MBDP algorithm.

The proof of the following theorem uses an idea introduced by [Szer et al., 2005]: A joint policy $\delta^T$ can be decomposed into a depth-$t$ policy tree $\delta^t$ and a *completion* of this policy tree $\Delta^{T-t}$ such that $\langle \delta^t, \Delta^{T-t} \rangle$ constitutes the complete policy vector $\delta^T$. The partial policy tree $\delta^t = \langle \delta_1^t, ..., \delta_n^t \rangle$ is the usual vector of local single-agent policies. The completion $\Delta^{T-t} = \langle \Delta_1^{T-t}, ..., \Delta_n^{T-t} \rangle$ is a vector of completions, one for each agent. After executing policy tree $\delta^t$ each agent will be at some leaf node in its policy tree as a result of a sequence of $t$ observations $\theta_i = (o_i^1, ..., o_i^t)$. The completion $\Delta^{T-t}$ must specify which policy tree to choose for each possible joint observation sequence $\theta$. Thus, $\Delta^{T-t}$ is a set of horizon-$T-t$ joint policy trees and $\Delta^{T-t}(\theta)$ specifies one particular joint policy tree for $\theta$. We can now define the value of a joint policy tree $\delta^T$ and an initial belief state $b^0$ as
$V(\delta^T, b^0) = V(\langle \delta^t, \Delta^{T-t} \rangle, b^0) = V(\delta^t, b^0) + V(\Delta^{T-t}|b^0, \delta^t)$.

**Theorem 1.** *For a given maxObs $< |O|$ and any belief state $b \in \Delta S$, the error of improved MBDP due to partial backups with selected observations on a horizon-$T$ problem, $|V(\delta^T, b) - V(\delta^T_{maxObs}, b)|$, is bounded by:*
$$\mu_T = T^2 \cdot (1 - \varepsilon) \cdot (R_{\max} - R_{\min})$$

**Proof:** Let $\Delta$ denote the optimal completion of a policy tree and $\widehat{\Delta}$ denote the completion computed by the improved MBDP algorithm. Let $\delta^T = \langle \delta^t, \widehat{\Delta}^{T-t} \rangle$ be the joint policy tree constructed by the improved MBDP algorithm. The proof is by induction on depth $T - t$ of the completion $\widehat{\Delta}^{T-t}$. Here the base case is $T - t = 2$, because for $T - t = 1$ the construction of the completion with horizon 1 does not yet require a backup.

**Base case ($T - t = 2$):** For $\langle \delta^{T-2}, \widehat{\Delta}^2 \rangle$ assume that $\delta^{T-2}$ is the optimal joint policy tree. Because $V(\delta^T, b^0) = V(\langle \delta^{T-2}, \widehat{\Delta}^2 \rangle, b^0) = V(\delta^{T-2}, b^0) + V(\widehat{\Delta}^2(\theta)|b^0, \delta^{T-2})$ we only have to look at the value of $\widehat{\Delta}^2$. Let $O_{\text{In}}$ denote the set of joint observations as identified by the algorithm which defines the sets of observations for each agent that are used for the partial backups. Analogously let $O_{\text{Out}}$ denote the set of observations that is not taken into consideration for the partial backup. The completion $\widehat{\Delta}^2$ is constructed using the result of the partial backup. Accordingly, when during execution the agents reach one of their policy tress in the completion and the next joint observation is $\in O_{\text{In}}$, no value loss occurs. However, if an observation from $O_{\text{Out}}$ is observed the policy tree might lead to a suboptimal action for the last step resulting in a maximum error of $(R_{\max} - R_{\min})$. Thus, for $\widehat{\Delta}^2$ the maximum value loss is bounded by $\mu_1 = (1 - \varepsilon) \cdot (R_{\max} - R_{\min})$.

**Inductive step ($t \to t+1$):** Now assume that $\mu_t = t^2 \cdot (1-\varepsilon)(R_{\max} - R_{\min})$. Let $\delta^T = \langle \delta^{T-(t+1)}, \widehat{\Delta}^{t+1} \rangle$. Then $V(\delta^T, b^0) = V(\langle \delta^{T-(t+1)}, \widehat{\Delta}^{t+1} \rangle, b^0) = V(\delta^{T-(t+1)}, b^0) + V(\widehat{\Delta}^{t+1}|b^0, \delta^{T-(t+1)})$. The value of the completion is defined as $V(\widehat{\Delta}^{t+1}|b^0, \delta^{T-(t+1)}) = \sum_{\theta \in \Theta} P(\theta|b^0, \delta^{T-(t+1)}) \cdot \sum_{s \in S} P(s|\theta) \cdot V(\widehat{\Delta}^{t+1}(\theta), s)$.

Thus we need only look at the last term:

$$V(\widehat{\Delta}^{t+1}(\theta), s) \tag{1}$$
$$= V(\hat{\delta}^1(\theta), s) + V(\widehat{\Delta}^t|s, \hat{\delta}^1(\theta)) \tag{2}$$
$$= V(\hat{\delta}^1(\theta), s) + \Big[ \sum_{\theta' \in \Theta} P(\theta'|s, \hat{\delta}^1(\theta))$$
$$\cdot \sum_{s' \in S} P(s'|\theta') V(\widehat{\Delta}^t(\theta'), s') \Big] \tag{3}$$
$$\geq V(\hat{\delta}^1(\theta), s) + \Big[ \sum_{\theta' \in \Theta} P(\theta'|s, \hat{\delta}^1(\theta))$$
$$\cdot \sum_{s' \in S} P(s'|\theta') V(\Delta^t(\theta'), s') - \mu_t \Big] \tag{4}$$
$$= V(\hat{\delta}^1(\theta), s) + V(\Delta^t|s, \hat{\delta}^1(\theta)) - \mu_t \tag{5}$$
$$\geq \varepsilon \cdot V(\Delta^{t+1}|s, \delta^{T-(t+1)}) + (1-\varepsilon)(t+1)R_{\min} - \mu_t \tag{6}$$
$$= \varepsilon \cdot V(\Delta^{t+1}|s, \delta^{T-(t+1)}) + (1-\varepsilon)(t+1)R_{\min}$$
$$+ (t+1)\mu_1 - (t+1)\mu_1 - \mu_t \tag{7}$$
$$= \varepsilon \cdot V(\Delta^{t+1}|s, \delta^{T-(t+1)})$$
$$+ (1-\varepsilon)(t+1)R_{\max} - (t+1)\mu_1 - \mu_t \tag{8}$$
$$\geq V(\Delta^{t+1}|s, \delta^{T-(t+1)}) - (t+1)^2 \mu_1 \tag{9}$$
$$= V(\Delta^{t+1}|s, \delta^{T-(t+1)}) - \mu_{t+1} \tag{10}$$

Step (2) follows from the definition of a decomposition. Note that $\hat{\delta}^1(\theta)$ denotes a one-step policy as computed by improved MBDP. In (4) we plugged in the inductive assumption. Note that in step (5) $\Delta^t$ is the optimal completion for the not necessarily optimal $\hat{\delta}^1(\theta)$. In (6) we use the fact that the sum of the probabilities taken into account for the partial backup is $\varepsilon$. Thus with probability $\varepsilon$ $\hat{\delta}^1$ leads to an optimal path and with probability $(1-\varepsilon)$ it leads to a suboptimal and potentially worst-case path. In (7) we add zero and (9) is based on the inequality: $(t+1)^2 > t^2 + t + 1$.

Thus, $\delta^T = \langle \delta^{T-t+1}, \widehat{\Delta}^{t+1} \rangle$ has at most error $\mu_{t+1}$. Accordingly, for all $T$, the error is bounded by $\mu_T$. This concludes the proof of the theorem. □

### 5.2 Convergence

**Corollary 1.** *Increasing $maxObs \to |O|$ the error bound $\mu_T = |V(\delta^T, b) - V(\delta^T_{maxObs}, b)|$ is strictly decreasing and reaches 0 in the limit.*

**Proof:** This result follows directly from Theorem 1. As $\varepsilon$ is strictly increasing in $maxObs$ the error bound is strictly decreasing in $maxObs$. When $maxObs$ reaches $|O|$ full backups instead of partial backups are performed in each step and thus the error bound is 0. □

### 5.3 Complexity Properties

**Theorem 2.** *For a fixed number of agents, the improved MBDP algorithm has a polynomial time and space complexity in all other input parameters.*

**Proof:** We have already shown that once the parameter $maxTrees$ is fixed, the time and space complexity of MBDP with regard to the problem horizon is polynomial [Seuken and Zilberstein, 2007]. The most costly step of MBDP is the full backup, which is exponential in the size of the observation space. The partial backup approximation in the improved MBDP algorithm is obviously polynomial in the observation space once $maxObs$ is fixed. Filling up the missing observation branches using local search at the end of every iteration can also be done in polynomial time: There are at most $|O|$ observations that can be connected to $maxTrees$ subtrees. Finding the most likely observations takes $|S| \cdot |O|^n$ operations which is also polynomial when $n$ is fixed. From Algorithm 1 it is obvious to see that the number of available actions as well as the size of the state space affect the complexity only polynomially. Thus, for a fixed number of agents we obtain a polynomial complexity in all other input parameters. □

Note that this complexity analysis only considers the case when $maxObs$ is fixed. It does not describe how the computational complexity scales with the desired accuracy (i.e. where $maxObs$ is variable). Unfortunately this analysis is not possible in general because how high $maxObs$ has to be set to reach a desired accuracy does not depend on the size of the input, i.e. the size of the observation space or other singleton parameters, but on all entries of the observation probability table. In particular, if every observation is possible in every state and very important in making the correct decisions then the solution value achieved from the approximate algorithm can be bad whenever $maxObs < |O|$. This shows that the error bound has primarily a theoretical significance and analyzing the trade-off between computational complexity and solution value can only be done experimentally.

## 6 Experiments

### 6.1 Existing Benchmark Problems

Most researchers working on DEC-POMDPs report performance results for the *multi-agent broadcast channel (MABC) problem* and the *multi-agent tiger problem*. The former problem has 2 agents, 4 states, 2 actions and 2 observations. The later problem has 2 agents, 2 states, 3 actions and 2 observations. Because partial backups are not necessary for an observation space of size 2, for these benchmark problems the improved MBDP algorithm leads to the same results as the original MBDP algorithm. The left part of Table 6 shows the results achieved by the MBDP algorithm on the MABC problem. More detailed results including running time analyses are reported in [Seuken and Zilberstein, 2007]. Shown are the solution values achieved for each horizon where a "-" indicates that the algorithm ran out of time or memory for this horizon. As the results show, MBDP reached the same or higher value than all existing algorithms while being able to solve horizons that were multiple orders of magnitude larger than any other algorithm could solve (up to 100,000 within 24h on the MABC problem). Even though no comparison with optimal solution algorithms is possible for the larger horizons, the solution values suggest that the algorithm achieves the optimal value for every horizon. This illustrates the need for a new, larger benchmark problem.

### 6.2 Cooperative Box-Pushing

The cooperative box-pushing problem is a well-known robotics problem introduced by [Kube and Zhang, 1997]. In this domain, two or more agents have to cooperate to move a big object (the box) that they could not move on their own. Even though the robots cannot communicate with each other, they have to achieve a certain degree of coordination to move the box at all.

We model this problem as a DEC-POMDP and introduce it as a new benchmark problem for algorithms that can handle larger observation spaces. Figure 2 shows a sample problem in a grid world domain.

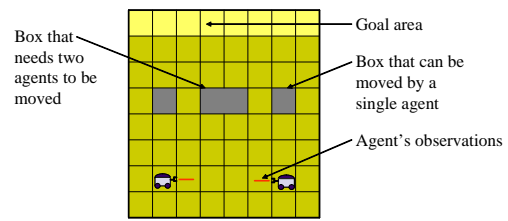

Figure 2: Box Pushing in the Grid World.

Table 1: Performance Comparison on the MABC Problem and the Box-Pushing Problem.

| Horizon | MABC Problem | | | | Cooperative Box-Pushing Problem | |
|---|---|---|---|---|---|---|
| | Optimal | PBDP | Random | **MBDP** | Random | **Improved MBDP** |
| 1 | 1.00 | 1.00 | 0.50 | **1.00** | −0.20 | **-0.20** |
| 2 | 2.00 | 2.00 | 1.00 | **2.00** | −0.06 | **17.60** |
| 3 | 2.99 | 2.99 | 1.49 | **2.99** | −0.75 | **65.68** |
| 4 | 3.89 | 3.89 | 1.99 | **3.89** | −1.69 | **97.91** |
| 5 | - | 4.70 | 2.47 | **4.79** | −2.67 | **100.64** |
| 6 | - | 5.69 | 2.96 | **5.69** | −3.73 | **117.63** |
| 7 | - | 6.59 | 3.45 | **6.59** | −4.84 | **133.12** |
| 8 | - | 7.40 | 3.93 | **7.49** | −5.97 | **182.92** |
| 9 | - | - | 4.41 | **8.39** | −7.12 | **186.95** |
| 10 | - | - | 4.90 | **9.29** | −8.30 | **189.32** |
| 20 | - | - | 9.73 | **18.29** | −20.46 | **415.25** |
| 50 | - | - | 24.23 | **45.29** | −57.94 | **1051.82** |
| 100 | - | - | 48.39 | **90.29** | −120.55 | **2112.05** |

This problem is particularly nice because it is very flexible in the sense that it can easily be modified to different sizes and problem complexities by varying the problem parameters. The system states of this problem are all tuples of possible positions of the two agents and the three boxes. Each agent has 4 actions: turn left, turn right, move forward and stay. If an agent is facing a box that it can move on its own and selects the action move forward, the box is pushed one grid cell into the direction of the agent's movement and the agent also moves one step forward. If the agent moves against a wall or against a larger box that it cannot push by itself it just stays in place. However, if both agents push against the large box at the same time, the large box moves by one grid cell as do the agents. To model an uncertain environment we assume that each agent's actions are only successful with probability 0.9 and with probability 0.1 it simply stays in place. After every time step each agent gets one out of 5 possible observations deterministically describing the situation of the environment in front of the agent: empty field, wall, other agent, small box, large box.

The reward function is designed such that the agents benefit from cooperation. Every time step the agents spend in the environment, they get a negative reward of -0.1 per agent. Per agent that bumps into a wall or a box it cannot move, they receive a penalty of -5. For each small box that reaches the goal area they get a reward of 10. If the agents cooperatively push the large box into the goal area they get a reward of 100.

As discussed before, due to the higher number of observations, none of the previous algorithms – including the MBDP algorithm – can be applied to this problem even for horizons less than 5. Only the improved MBDP algorithm can tackle this problem making use of the partial backups. The right part of Table 6 shows initial performance results with $maxObs = 3$ and $maxTrees = 3$ for the small problem depicted in Figure 3. Note that this small problem already has 100 states, 4 actions and 5 observations. The problem starts as shown in Figure 3. Upon reaching one of the goal states the system state transitions back into the start state. In our experiments the running time scales linearly with the horizon. The algorithm needs approx. 30 seconds per level of the horizon with $maxObs = 2$ and 500 seconds with $maxObs = 3$. At the moment our experimental results are only preliminary and should not be used for comparison yet. We will soon report more detailed experimental results that will also highlight the trade-off between computational complexity (running time) and solution quality.

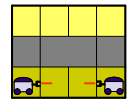

Figure 3: Small Grid World Problem.

## 7 Conclusion

The original MBDP algorithm was the first approximate algorithm that could be applied to DEC-POMDPs with non-trivial horizons. But the algorithm still suffered from exponential complexity in the size of the observation set. Our main contribution in this paper is the introduction of an improved version of MBDP that reduces the complexity with respect to the observation set from exponential to polynomial. This can be accomplished because for most belief states, many observations are highly unlikely or even impossible. Thus, performing a partial backup with only the most likely observations is sufficient to obtain high-quality solutions. We have also presented a theoretical analysis of the improved MBDP algorithm, proving worst-case bounds on the approximation error introduced by the partial backup approximation. We have shown a quadratic relationship between the error bound and the horizon which is interesting because intuitively one might think that the error accumulates exponentially. Furthermore, we have shown that when $maxObs$ is increased, the error bound decreases and converges to 0 in the limit. Because the existing benchmark problems are too simple to test the improved algorithm, we created a new benchmark problem called Cooperative Box Pushing that can be solved currently only by the improved MBDP algorithm. Our experimental results show that despite the double exponen-tial complexity of DEC-POMDPs, good solutions for reasonably hard problems can be computed efficiently.

One important remaining challenge is deriving a complete error bound for the improved MBDP algorithm. So far, we have only bounded the error introduced due to the partial backups with a selected set of observations. But we have not said anything about the error introduced due to the selection of policy trees based on belief space sampling. It turns out that bounding the error in this direction is very challenging. Successful error-bounding techniques for POMDPs have been introduced in [Pineau et al., 2003] and [Smith and Simmons, 2005]. They show that if the belief space of the POMDP can be sampled densely enough, the error introduced by their approximation algorithm can be minimized to any $\varepsilon > 0$. Unfortunately, for DEC-POMDPs this technique does not work because an optimal policy is not a mapping from belief states to actions. Instead, each agent has to base its decision on a multi-agent belief state, that is, a distribution over systems states and over the policies of the other agents. An algorithm that aims to approximate the optimal solution within an error bound must also, at least partly, base its decisions on the multi-agent belief state which it can infer from its sequence of observations during execution time. However, because we are constructing our policies bottom-up, a distribution over the other agent's policies is not available before the policies are fixed. This presents a difficulty that precludes a complete non-trivial error bound to be established. Furthermore, [Rabinovich et al., 2003] have shown that even $\varepsilon$-approximations for DEC-POMDPs are NEXP-hard to find. Thus, even if one finds an approximate algorithm for DEC-POMDPs with a complete error bound, the algorithm would at least have an exponential and most likely a double exponential complexity in the bound itself. Instead of pre-computing an error bound, one could collect information during the execution of an approximate algorithm to find an on-line bound. In future work we will try to establish such an on-line bound and to evaluate more rigorously the performance of the improved MBDP algorithm.

## Acknowledgments

We thank Marek Petrik for helpful discussions on this paper. This work was supported in part by the Air Force Office of Scientific Research under Award No. FA9550-05-1-0254 and by the National Science Foundation under Grants No. IIS-0219606 and IIS-0535061.